\documentclass{article}
\usepackage{ijcai25}

\usepackage{times}
\usepackage{soul}
\usepackage{url}
\usepackage[hidelinks]{hyperref}
\usepackage[utf8]{inputenc}
\usepackage[small]{caption}
\usepackage{graphicx}
\usepackage{amsmath}
\usepackage{amsthm}
\usepackage{booktabs}
\usepackage{algorithm}
\usepackage{algorithmic}
\usepackage[switch]{lineno}

\usepackage{graphicx} % Required for inserting images
\usepackage[backend=biber,style=alphabetic,]{biblatex}

\usepackage{comment}
\usepackage{rotating}
\usepackage{array}    % For better column formatting
\usepackage{multirow}
\usepackage{tikz}
\usepackage{makecell} % Add this in your preamble
\usepackage{adjustbox}

\title{\textsc{ParBalans}: Parallel Multi-Armed Bandits-based \\Adaptive Large Neighborhood Search}

\author{
Alican Yilmaz$^{1,2}$\and
Junyang Cai$^3$\and
Serdar Kad{\i}o\u{g}lu$^{2,4}$\and
Bistra Dilkina$^3$
\affiliations
$^1$Department of Mechanical \& Industrial Engineering, Northeastern University\\
$^2$AI Center of Excellence, Fidelity Investments\\
$^3$Department of Computer Science, University of Southern California\\
$^4$Department of Computer Science, Brown University\\
\emails
yilmaz.al@northeastern.edu,
caijunya@usc.edu,
dilkina@usc.edu,
serdark@cs.brown.edu
}

\begin{document}

\maketitle

\begin{abstract}
    Solving Mixed-Integer Programming (MIP) problems often requires substantial computational resources due to their combinatorial nature. Parallelization has emerged as a critical strategy to accelerate solution times and enhance scalability to tackle large, complex instances.
    This paper investigates the parallelization capabilities of \textsc{Balans}, a recently proposed multi-armed bandits-based adaptive large neighborhood search for MIPs. 
    % a recently proposed adaptive meta-solver for MIPs that incorporates online learning. 
    While \textsc{Balans}'s modular architecture inherently supports parallel exploration of diverse parameter configurations, this potential has not been thoroughly examined. To address this gap, we introduce \textsc{ParBalans}, an extension that leverages both solver-level and algorithmic-level parallelism to improve performance on challenging MIP instances. Our experimental results demonstrate that \textsc{ParBalans} exhibits competitive performance compared to the state-of-the-art commercial solver Gurobi, particularly on hard optimization benchmarks.
\end{abstract}

\section{Introduction}
\label{sec:intro}

Mixed-integer programming (MIP) is a versatile framework for expressing a wide range of combinatorial optimization problems. For solving real-world instances, exact solvers often require prohibitive time since solving MIPs to global optimality is NP-hard.  Therefore, practitioners focus on rapidly obtaining high-quality feasible solutions, typically through meta-heuristics such as large neighborhood search (LNS)~\cite{10.1007/3-540-49481-2_30}.

Recently, Cai et al.~\cite{balans,balansarxiv} proposed \textsc{Balans}, an online adaptive meta-solver that couples LNS with a multi-armed-bandit controller to adaptively select among diverse neighborhoods. \textsc{Balans} achieves competitive performance versus state-of-the-art solvers such as \textsc{SCIP}~\cite{bolusani2024scip} and \textsc{Gurobi}~\cite{gurobi}. Moreover, \textsc{Balans} is as a highly configurable, MIP solver agnostic, open-source software with a one-liner installation\footnote{pip install balans}$^{,\thinspace}$\footnote{\url{https://github.com/skadio/balans}}.

\textsc{Balans} is an integration technology that offers a modular architecture to leverage best-in-class open-source software dedicated to their specific domain. We leverage  \textsc{MABwiser}~\cite{DBLP:conf/ictai/StrongKK19,DBLP:journals/ijait/StrongKK21} for bandits, \textsc{ALNS} library~\cite{Wouda_Lan_ALNS_2023} for adaptive large-neighborhood search, and \textsc{SCIP}~\cite{bolusani2024scip} and \textsc{Gurobi}~\cite{gurobi} for MIP solving. Future advances in these distinct fields, realized independently within each software, propagate to our meta-solver with compounding effects. This unique integration makes \textsc{Balans} highly configurable. Beyond the single default configuration used in~\cite{balans}, more configurations are available through  parameterization of ALNS and MABWiser. The original study explores a single default  configurations, kept identical across different problem domains, instances, and solvers, to demonstrate the robustness of the main algorithm without any tuning, thus leaving potential performance untapped. Given that \textsc{Balans} is highly-configurable and  exposes a large hyperparameter space, different configuration settings can yield further performance boost as in instance-specific algorithm configuration~\cite{DBLP:conf/aaai/KadiogluMS12,DBLP:conf/cp/KadiogluMSSS11,DBLP:conf/ecai/KadiogluMST10,dash}.

\medskip
In this paper, we introduce \textsc{ParBalans}\footnote{\url{https://github.com/skadio/balans\#quick-start---parbalans}}, a parallel variant of \textsc{Balans} designed to exploit the large configuration space at the intersection of adaptive search, meta-heuristics, multi-armed bandits, and mixed-integer programming. Our contributions are threefold:

\begin{itemize}
    \item Empirical motivation for parallelism. By evaluating hundreds of  \textsc{Balans} configurations, run in parallel, on difficult benchmark instances across problem types, we show that no single setting dominates all problems, underscoring the value of running multiple configurations concurrently.
    \item Randomized configuration generation. We devise a lightweight algorithm that samples diverse, high-potential parameter sets.
    \item Extensive computational study. We benchmark \textsc{ParBalans} on larger, harder, and industrial MIP instances, comparing it with the default single-thread and parallel Gurobi. \textsc{ParBalans} consistently obtains competitive results, demonstrating that a simple parallel exploration of hyperparameter space can rival, and sometimes surpass, sophisticated parallel branch-and-bound.
\end{itemize}

Overall, our empirical findings suggest that parallel meta-solvers, as demonstrated by \textsc{ParBalans}, is a promising complement to existing parallel exact-solver strategies, opening new avenues for scalable MIP solving. 

\medskip
Let us start by the sources of parallelism, the algorithm configuration space, how we run simulations, and then present our experimental results.

\section{Sources of Parallelism}
\label{sec:parallelism}

\textsc{ParBalans} can leverage modern multi-core architectures at two layers of parallelization strategy: 

\begin{enumerate}
    \item At the solver layer, we utilize Gurobi's native multi-threading capabilities by tuning the \texttt{Threads} parameter (or any other MIP solver that can take advantage of multi-threading). 
    \item At the algorithmic layer, we run independent \textsc{Balans} workers, each initialized with a different configuration. These two layers of parallelism compose and complement each other. 
\end{enumerate}

Overall, parallelism is the product of the number of processes and the number of threads per process.

\begin{comment}
To maximize hardware utilization, we adopt a two-tier core allocation scheme. At the process level, a \texttt{multiprocessing.Pool} evenly distributes BALANS instances—each with its own \texttt{Gurobi Env()} and \texttt{Model()}—across $P$ Python processes. At the solver level, each process sets Gurobi's \texttt{Threads} parameter to $T$, enabling parallelization of the branch-and-bound procedure across $T$ cores. To prevent thread oversubscription, we empirically choose the number of processes as $P \approx \text{\# of physical cores} \div T$.
\end{comment}

\section{Configuration Space}
\label{sec:config}

\begin{comment}
paragraph 1: pool of options
\end{comment}

The \textsc{Balans} framework leverages a diverse pool of configurations to explore a broad range of hyper-parameter space. Each configuration is defined by a unique set of parameters, which includes categorical design choices among i) destroy operators, ii) accept criterion, and iii) learning policy, and numerical hyper-parameters such as destroy percentages. A comprehensive summary of all configurable parameters, their types, and valid ranges is provided in Appendix A. 

In our \textsc{ParBalans} experiments we consider 180 different configurations which neatly fits within Amazon EC2 Instances with 192 cores. To generate these configurations, we employ a lightweight random sampling procedure that selects each parameter independently from its predefined pool, while respecting inter-parameter dependencies and design constraints. For example, if the Thompson Sampling learning policy is selected, the reward value pool is restricted to binary vectors such as 
[1,1,1,0] or [1,1,1,0]; similarly, if the Proximity Search destroy operator is chosen, its destroy percentage is sampled from 
[5,10,15,20,30]. 

\begin{comment}
Our configuration design includes specialized parameter settings for certain components. For instance, the proximity search destroy operator uses a percentage pool of $[5,10,15,20,30]$, while the Thompson Sampling learning policy restricts reward value pools to $[1,1,1,0]$ or $[1,1,0,0]$. 

paragraph 2: random config generator

To generate the configuration pool, we use a lightweight random sampling algorithm. Each parameter is sampled independently from its respective pool, while respecting inter-parameter dependencies and design choices. For example, if Thompson Sampling is selected, the reward pool is adjusted accordingly; similarly, the proximity search operator requires a distinct percentage pool.
\end{comment}
Algorithm 1 formalizes the destroy arm selection procedure. The total number of destroy operators (i.e., arms) in each configuration is selected from the range $[4,16]$, allowing us to explore the trade-off between exploration and exploitation. A larger number of arms may hinder \textsc{Balans}'s ability to learn effectively during the search, as each arm may not be utilized frequently enough for learning. Conversely, fewer arms may limit the potential benefits of leveraging diverse destroy operators. In our previos work, we highlight the importance of operator diversity in achieving superior performance over single-neighborhood LNS approaches~\cite{balans}.

To ensure diversity and avoid redundant configurations, we enforce a constraint that requires at least one operator from each destroy category, when applicable. This guarantees balanced representation across operator types. 

\begin{algorithm}[t]
\caption{A simple procedure to generate different destroy operator configurations.}
\begin{algorithmic}[1]
\STATE \textbf{Input:} Dictionary of destroy categories:
\STATE \hspace{1em} \texttt{crossover}: [c]
\STATE \hspace{1em} \texttt{mutation}: [m\_10, m\_20, m\_30, m\_40, m\_50]
\STATE \hspace{1em} \texttt{local\_branch}: [lb\_10, lb\_20, lb\_30, lb\_40, lb\_50]
\STATE \hspace{1em} \texttt{proximity}: [p\_05, p\_10, p\_15, p\_20, p\_30]
\STATE \hspace{1em} \texttt{rens}: [r\_10, r\_20, r\_30, r\_40, r\_50]
\STATE \hspace{1em} \texttt{rins}: [ri\_10, ri\_20, ri\_30, ri\_40, ri\_50]
\STATE \textbf{Output:} List of selected destroy operators
\STATE
\STATE Randomly select number of operators $N \in [4, 16]$
\IF{$N \geq 6$}
    \STATE Select one operator randomly from each category
    \STATE Select additional $N - 6$ operators randomly from the remaining pool
\ELSE
    \STATE Randomly select $N$ categories
    \STATE Select one operator randomly from each selected category
\ENDIF
\STATE Return the list of selected operators
\end{algorithmic}
\end{algorithm}

\section{Simulations}
\label{sec:simulations}
\begin{comment}
paragraph 1: Explain generally how \textsc{ParBalans} is simulated and how pb1-2-4-8 graphs are obtained(randomly selecting N balans for parbalans N 1000 times) 
\end{comment}

Given the configuration space in Section~\ref{sec:config}, we construct a pool of 180 unique  \textsc{Balans} configurations guided by the Algorithm 1.

Next, we perform 1,000 simulation runs, each defined by a randomly sampled subset of $N$ configurations from this pool. To ensure reproducibility, we seed each process consistently during the simulation.

In each run, \textsc{ParBalans} $N$ is executed on every test instance for one hour. The $N$ parallel  \textsc{Balans} workers periodically communicate their current best primal gaps. At any time $t$, the overall primal gap is defined as the minimum among these $N$ values. The final reported gap corresponds to the best value observed across all workers by the end of the run.

\begin{comment}
paragraph 2: Explain how lower bounds are simulated
\end{comment}

After completing \textsc{ParBalans} $N$ on all instances, we record the average primal gap and the identifiers of the selected configurations. Aggregating results across all 1,000 runs allows us to estimate the expected performance distribution of \textsc{ParBalans} $N$ and identify the single best-performing configuration subset.

\begin{comment}
paragraph 3: explain how multi-thread gurobi parbalans are simulated. Explain the machine limitation and tell that those X conifgs are chosen randomly from the 180 pool. 
\end{comment}

As an initial step, we set the Gurobi \texttt{Threads} parameter to 1 for each \textsc{Balans} instance, allowing all 180 configurations from the pool to be executed fully in parallel during a single run. We then conduct simulations for \textsc{ParBalans} with $N = 2, 4, 8, 16, 32, 64, 128$, where $N$ denotes the number of concurrently running  \textsc{Balans} processes.
\begin{comment}
correct this paragraph after finishing the Q2. Threads numbers might change depending on what i put into report. 
\end{comment}

To explore higher levels of solver-level parallelism, we increase the value of \texttt{Threads} to 4, 8, and 16, which correspondingly limits the number of  \textsc{Balans} processes per run to 45, 20, and 10, respectively. Under these settings, \textsc{ParBalans} is executed with $N = 2, 4, \ldots, 32$ for \texttt{Threads} = 4, with $N = 2, 4, \ldots, 16$ for \texttt{Threads} = 8, and with $N = 2, 4, 8$ for \texttt{Threads} = 16. 

\begin{figure*}[t]
    \centering
    \includegraphics[width=0.85\textwidth]{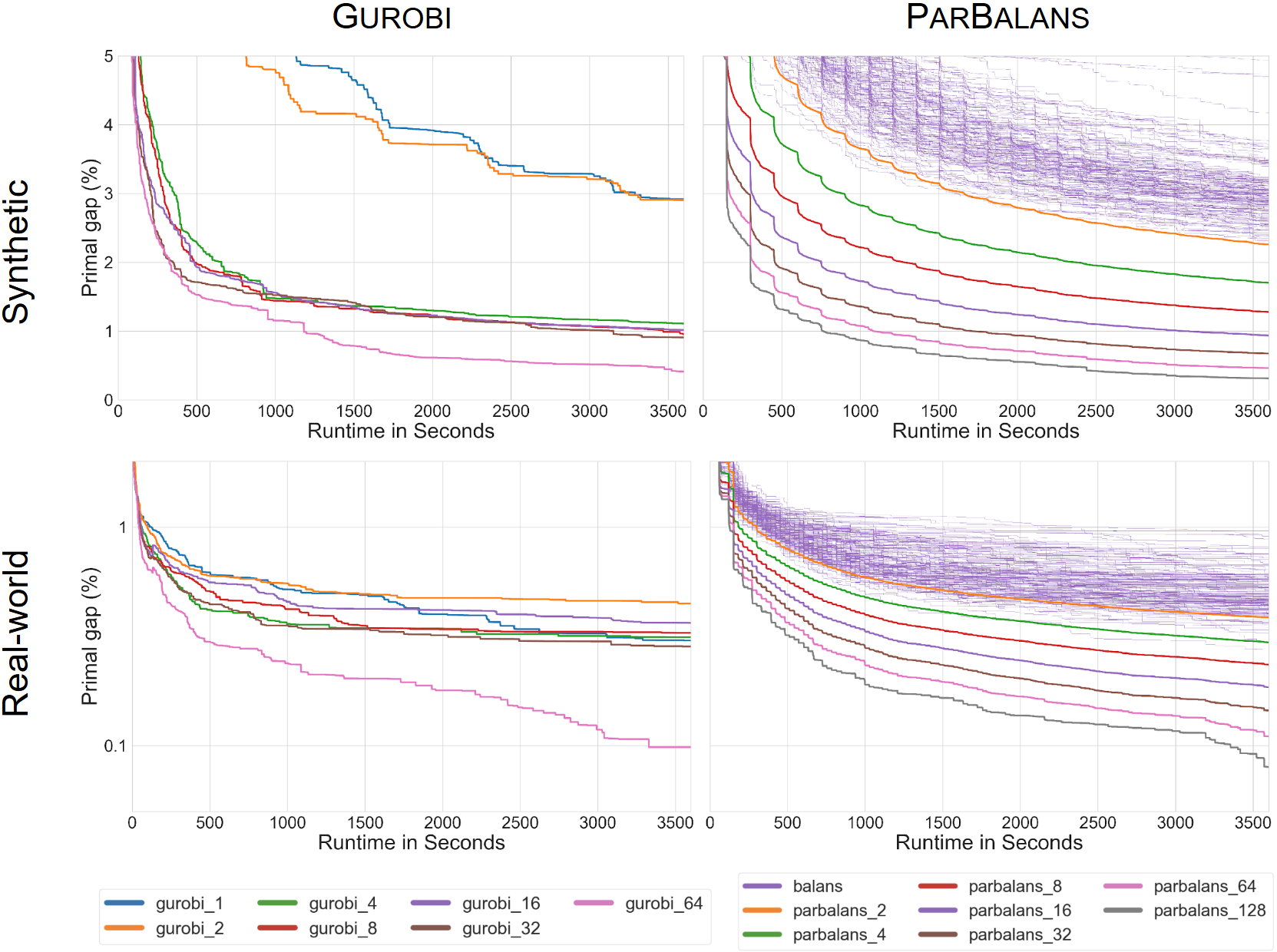}
    \caption{Primal gap (lower is better) as a function of time, averaged over instances from D-MIPLIB Synthetic (top) and D-MIPLIB Real-world (bottom). The plots compare the performance of \textsc{Gurobi} and \textsc{ParBalans}. The left column shows \textsc{Gurobi} with thread counts of 1, 2, 4, 8, 16, 32, and 64; the right column shows \textsc{ParBalans} with 1, 2, 4, 8, 16, 32, 64 and 128 configurations. Each colored line represents a different level of parallelization. Purple lines indicate the performance of individual \textsc{Balans} runs (no parallelization). The bottom row uses a logarithmic scale for the y-axis to enhance visibility.}
    \label{fig:q1}
\end{figure*}

This controlled simulation design ensures that the total number of logical threads (i.e., $N \times \texttt{Threads}$) never exceeds the available vCPUs, in our case Amazon EC2 Instances with 192 cores, thereby avoiding resource contention. It also enables a fair assessment of the trade-off between algorithm-level and solver-level parallelism.

\begin{comment}

Simulation also enables us to estimate both the best-performing (lower-bound) and worst-performing (upper-bound) configurations among different \textsc{ParBalans} $N$ instances. Not surprisingly, the variance (hence confidence intervals) in the performance of \textsc{ParBalans} $N$, decreases as $N$ increases (e.g., $N = 1, 2, 4, \ldots$).
\end{comment}

\medskip
\noindent \textbf{Notation:} We use the notation \textsc{ParBalans}-$N$(\textsc{Gurobi}-$T$) to denote a setup where $N$  \textsc{Balans} instances are executed in parallel, each using a \textsc{Gurobi} solver with $T$ threads. This results in a total parallelization of $N \times T$. For example, \textit{ParBalans-16(Gurobi-4)} refers to 16 parallel \textsc{Balans} processes, each utilizing 4 \textsc{Gurobi} threads.

\section{Experiments}
To quantitatively analyze the benefits of \textsc{Balans} parallelization, we run a series of experiments across a diverse set of problem instances. Below, we detail the experimental setup followed by a comprehensive analysis of the experiment results.

\subsection{Setup: Benchmarks, Metrics, Machines}

\begin{comment}
Paragraph 1: Explain why benchmark selection is important and we tested both hard and real life problems. 
\end{comment}

We select datasets from Distributional MIPLIB (D-MIPLIB), a standardized library of MIP problems categorized by different difficulty levels~\cite{huang2024distributional}. 

\medskip

To compare performance differences between parallel \textsc{Gurobi} and \textsc{ParBalans}, we deliberately focused on the hardest problem instances in D-MIPLIB. Notably, for small, medium, and even some real-world problems, both \textsc{Gurobi} and \textsc{ParBalans} often converge to the global optimum quickly with minimal parallelization, making it difficult to distinguish their performance or assess the impact of scaling. 

\begin{comment}
Paragraph 2: Explain the datasets. miplibd h etc.
\end{comment}

\begin{comment}
Mvc and mk are removed since it gives 0 gap. In the plots and summary tables, we grouped sc, mis, gisp and ca as synthetic ( total of 40 instances) and mmcn, srpn ( total of 20 instances) as "real-world". 
\end{comment}

\medskip
\noindent \textbf{Benchmarks:} Specifically, we consider a subset of synthetic and real-world datasets from D-MIPLIB. For synthetic data, we randomly select 10 instances each from Combinatorial Auctions (CA), Set Covering (SC), Maximum Independent Set (MIS), and General Independent Set Problems (GISP), yielding 40 instances in total. For real-world data, we select 10 instances each from the Middle-Mile Consolidation Network (MMCN) and Seismic-Resilient Pipe Network Planning problems (SRPN), resulting in 20 real-world instances. All instances are selected from the highest difficulty levels.

\begin{table*}[t]
\centering
\begin{adjustbox}{max width=\textwidth}
\begin{tabular}{c|cc|cc||cc|cc}
\toprule
\multirow{2}{*}{Degree of Parallelism} 
& \multicolumn{4}{c||}{\textbf{Synthetic}} 
& \multicolumn{4}{c}{\textbf{Real-world}} \\
& \multicolumn{2}{c|}{\textsc{Gurobi}} & \multicolumn{2}{c||}{\textsc{Parbalans}}
& \multicolumn{2}{c|}{\textsc{Gurobi}} & \multicolumn{2}{c}{\textsc{Parbalans}} \\
\cmidrule(lr){2-3} \cmidrule(lr){4-5} \cmidrule(lr){6-7} \cmidrule(lr){8-9}
& PG (\%) & PI & PG (\%) & PI & PG (\%) & PI & PG (\%) & PI \\
\midrule
2   & 2.91±2.91 & 128.01±135.25 & \textbf{2.26±1.70} & 99.16±60.29 & \textbf{0.45±0.44} & 16.15±15.94 & 0.39±0.38 & 16.49±15.64 \\
4   & \textbf{1.11±1.05} & 45.60±39.61 & 1.70±1.46 & 76.29±54.99 & \textbf{0.31±0.38} & 11.22±12.69 & 0.30±0.30 & 13.18±13.05 \\
8   & \textbf{0.94±1.04} & 42.48±38.53 & 1.28±1.23 & 58.90±48.02 & 0.33±0.36 & 12.03±12.84 & \textbf{0.24±0.25} & 10.82±11.01 \\
16  & 1.01±1.47 & 43.01±48.28 & \textbf{0.94±1.06} & 44.97±41.86 & 0.36±0.48 & 13.91±16.03 & \textbf{0.19±0.20} & 8.93±9.32 \\
32  & 0.90±1.58 & 41.01±49.04 & \textbf{0.67±0.91} & 34.44±36.30 & 0.28±0.30 & 10.79±10.38 & \textbf{0.14±0.17} & 7.45±8.01 \\
64  & \textbf{0.41±0.54} & 26.01±19.90 & 0.46±0.76 & 26.35±31.89 & \textbf{0.10±0.19} & 6.00±7.08   & 0.11±0.14 & 6.23±6.94 \\
128 & N/A       & N/A           & \textbf{0.31±0.64} & 20.58±28.59 & N/A       & N/A         & \textbf{0.08±0.12} & 5.23±6.09 \\
\bottomrule
\end{tabular}
\end{adjustbox}
\caption{Performance comparison of Parallel \textsc{Gurobi} and \textsc{ParBalans} on synthetic (left) and real-world (right) instances. 
Each entry reports the average Primal Gap (PG, in \%) and Primal Integral (PI) at a 1-hour time limit. 
The best-performing method at each thread count is bolded. 
Lower values indicate better performance. 
To ensure fairness, PI is computed from minute 6 to hour 1 to mitigate early-stage solver advantages.}
\label{tab:q1}
\end{table*}

\medskip
\noindent \textbf{Evaluation Metrics:} We primarily use two performance metrics for evaluation, following~\cite{balans}: \textit{primal gap} and \textit{primal integral}. The \textit{primal gap}  measures the relative difference between the current primal bound $x$ and the best known objective value $x^*$, and is defined as:
\[
\frac{|x - x^*|}{\max(|x^*|, \epsilon)}
\]

The \textit{primal integral} quantifies the convergence behavior of the primal bound over time by integrating the primal gap from the start of the solving process up to time $t$. Formally, at time $t$, the primal integral is defined as the integral of the primal gap over the interval $[0, t]$. 
\begin{comment}
 In our experiments, to mitigate the unrealistic head-start advantage observed in \textsc{ParBalans}, we compute the primal integrals starting from minute 6 up to hour 1 for both Gurobi and \textsc{ParBalans}.   
\end{comment}

\medskip
\noindent \textbf{Machines:} We conduct experiments on Amazon SageMaker using an ml.c7i.48xlarge instance, which provides 192 vCPUs, and 384 GiB of memory. This high-performance compute environment enables the efficient evaluation of \textsc{ParBalans} and Parallel Gurobi across all problem instances.

\subsection{Numerical Results}

In our experimental evaluation, our main goal is to understand how algorithmic-level parallelism in \textsc{ParBalans} scales in comparison to direct solver-level parallelism in Gurobi. Specifically, the research questions are: 

\begin{itemize}
    \item \textbf{Q1:} How does \textsc{ParBalans}, using $N$ single-threaded processes (i.e., \textsc{Gurobi} with \texttt{Threads} = 1), perform relative to \textsc{Gurobi} with $N$ threads?
    
    \item \textbf{Q2:} How does \textsc{ParBalans}, using $N$ processes each running \textsc{Gurobi} with $T$ threads, perform relative to \textsc{Gurobi} with $N \times T$ threads?\
\end{itemize}

To answer these, we conduct simulations with varying parallelism, evaluating \textsc{ParBalans} with $N = 2, 4, 8, 16, 32, 64, 128$ and \textsc{Gurobi} with multiple threads. 

\begin{table}[t]
\centering
\begin{adjustbox}{max width=\columnwidth}
\begin{tabular}{@{}c c c c c@{}}
\toprule
& \makecell{\textbf{Degree of} \\ \textbf{Parallelism}} & \makecell{\textbf{{\textsc{Gurobi}}} \\ \textbf{Gap (\%)}} & \makecell{\textbf{\textsc{ParBalans}} \\ \textbf{Avg. Gap (\%)}} & \makecell{\textbf{\textsc{ParBalans}} \\ \textbf{Best Gap (\%)}} \\
\midrule
\multirow{7}{*}{\rotatebox[origin=c]{90}{\textbf{Synthetic}}} 
& 2   & 2.91 & 2.26 & \textbf{1.65} \\
& 4   & 1.11 & 1.70 & \textbf{1.11} \\
& 8   & 0.94 & 1.28 & \textbf{0.79} \\
& 16  & 1.01 & 0.94 & \textbf{0.52} \\
& 32  & 0.90 & 0.67 & \textbf{0.42} \\
& 64  & 0.41 & 0.46 & \textbf{0.31} \\
& 128 & --   & 0.31 & \textbf{0.27} \\
\midrule
\multirow{7}{*}{\rotatebox[origin=c]{90}{\textbf{Real-world}}} 
& 2   & 0.45 & 0.39 & \textbf{0.23} \\
& 4   & 0.31 & 0.30 & \textbf{0.18} \\
& 8   & 0.33 & 0.24 & \textbf{0.14} \\
& 16  & 0.36 & 0.19 & \textbf{0.12} \\
& 32  & 0.28 & 0.14 & \textbf{0.09} \\
& 64  & 0.10 & 0.11 & \textbf{0.07} \\
& 128 & --   & 0.08 & \textbf{0.07} \\
\bottomrule
\end{tabular}
\end{adjustbox}
\caption{Primal gap comparison across parallelization levels for \textsc{Gurobi} and \textsc{ParBalans} best configuration performance.}
\label{tab:best_config}
\end{table}

\begin{figure*}[t]
    \centering
    \includegraphics[width=0.85\textwidth]{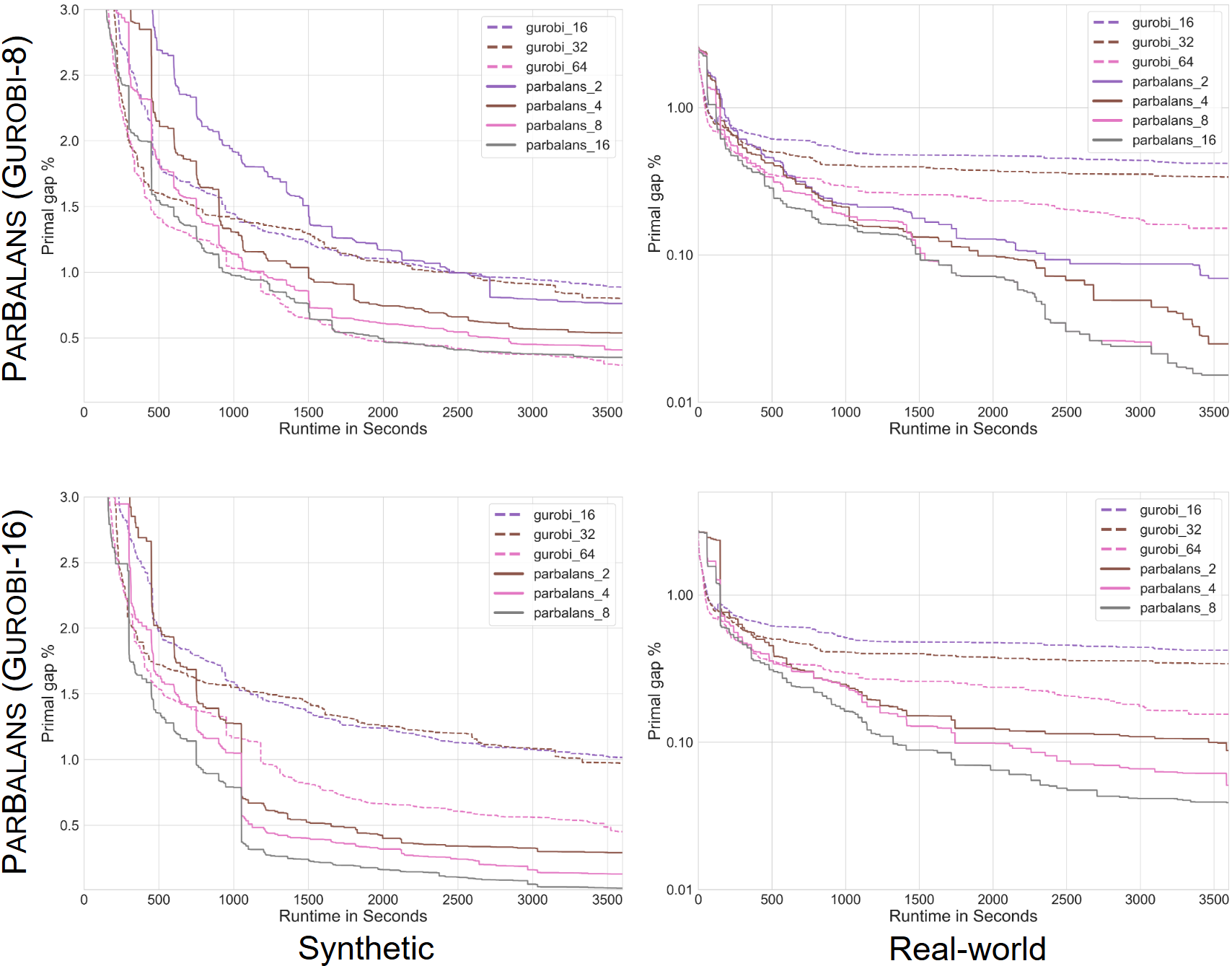}
    \caption{Performance comparison between \textsc{ParBalans} and parallel \textsc{Gurobi} across varying levels of parallelization on Synthetic (left) and Real-world (right) instance sets. This analysis explores how \textsc{ParBalans} scales when \textsc{Gurobi}’s internal thread count is set to 8 and 16. Dashed lines represent different levels of parallelism for \textsc{Gurobi}, while solid lines indicate varying degrees of parallelization for \textsc{ParBalans}. Line colors are matched across both solvers to represent equivalent concurrency levels for direct comparison.}
    \label{fig:q2}
\end{figure*}

\subsubsection{Q1: \textsc{ParBalans} $N$ (\textsc{Gurobi} 1) vs. \textsc{Gurobi} $N$}
Figure~\ref{fig:q1} shows that parallelization significantly enhances performance of \textsc{Balans} across both synthetic and real world D-MIPLIB instance sets. 

For synthetic instances (top row), on average, \textsc{Gurobi} outperforms \textsc{ParBalans} at lower parallelization levels, but the performance gap narrows as parallelization increases. 

For real-world instances (bottom row), \textsc{ParBalans} outperforms \textsc{Gurobi} at higher parallelization levels. Note that, \textsc{Gurobi}'s performance does not always improve with more threads, whereas \textsc{ParBalans} shows consistent gains with increased parallelization. This is likely due to the communicative nature of \textsc{Balans} configurations: different configurations excel on different instances, and no single configuration dominates across all the instances. This diversity is effectively leveraged through parallel execution.

\medskip
Table~\ref{tab:q1} presents the average primal gaps (\%) and primal integrals of both \textsc{Gurobi} and \textsc{ParBalans} after one hour, evaluated across two instance sets. It illustrates how performance varies across instances.

\begin{comment}
    this is not exactly # of threads for parbalans. is it still fine to name it like this? For now i named the column as "Degree of parallelism" ?
\end{comment}

\medskip
Table~\ref{tab:best_config} includes the best-performing \textsc{ParBalans} out of 1000 simulations to capture the full performance spectrum. Note that each sampled configuration remains fixed across all instances, and best-performing configurations selected based on their average performance across all instances. These results suggest that with careful configuration design, \textsc{ParBalans} performance could be further improved. \\

\begin{figure}[t]
    \raggedright
    \includegraphics[width=\columnwidth]{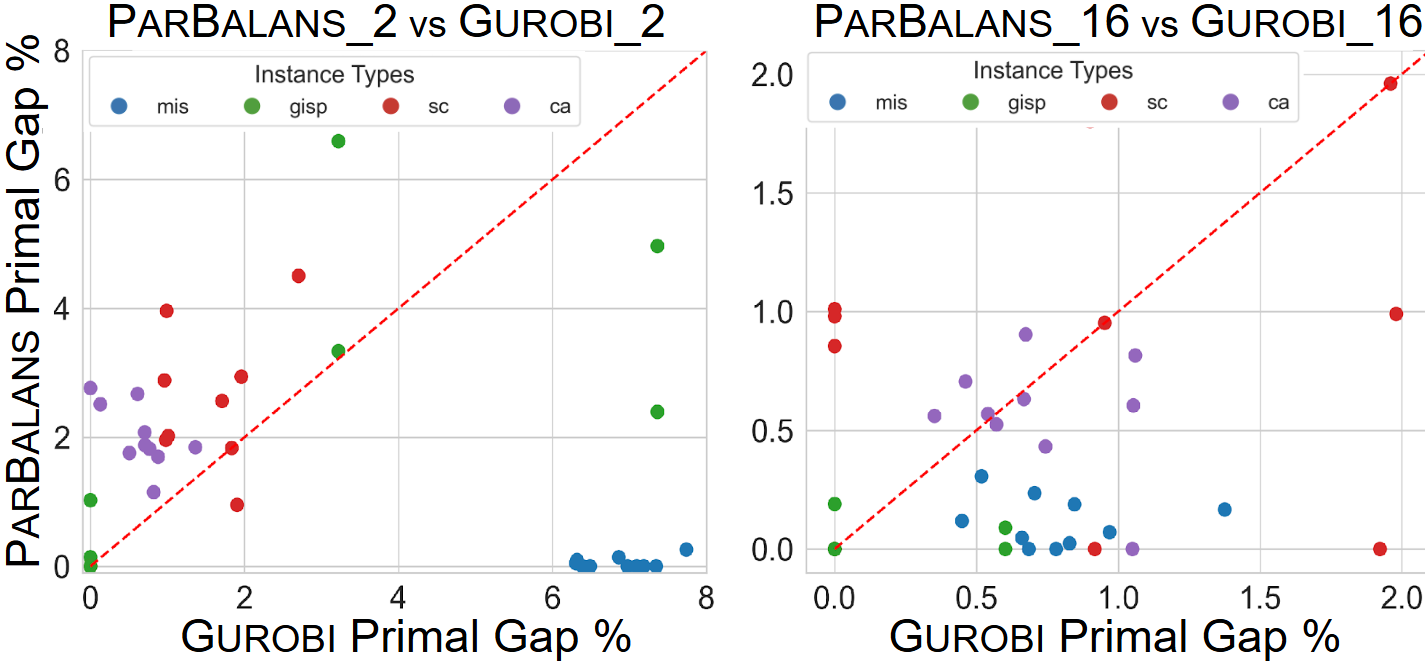}
    \caption{Primal gap between \textsc{ParBalans}-N (y-axis) vs \textsc{Gurobi}-N (x-axis) on Synthetic Instance Sets.}
    \label{fig:per_instance}
\end{figure}

Figure~\ref{fig:per_instance} presents a comparison of the primal gap of \textsc{ParBalans} and \textsc{Gurobi} on a per-instance basis, with points color-coded by instance set. Points below the diagonal line indicate that \textsc{ParBalans} outperforms \textsc{Gurobi} on that instance. For a fair comparison, each subplot compares equivalent levels of parallelization, e.g., \textsc{ParBalans}-2 vs. \textsc{Gurobi}-2.

As shown in Figure~\ref{fig:per_instance}, at lower levels of parallelization, \textsc{Gurobi} outperforms \textsc{ParBalans} on the majority of instances. However, as the degree of parallelization increases, \textsc{ParBalans} begins to outperform \textsc{Gurobi}, overall. This trend is particularly evident in the CA  problem set (purple): while \textsc{Gurobi}-2 consistently outperforms \textsc{ParBalans}-2, \textsc{ParBalans} surpasses \textsc{Gurobi} as N increases to 16. 

\begin{comment}
Another notable observation is that both \textsc{ParBalans} and Gurobi achieve a primal gap of zero at most parallelization levels for SRPN instance set, suggesting that both solvers frequently reach the global optimum.
\end{comment}

\begin{comment}
Should i plot instead win-lose of "best-parbalans vs gurobi ? 

When comparing Gurobi to the best-performing \textsc{ParBalans}-N configuration, the success rate improves substantially. We refer the reader to the Appendix for a detailed comparison between “Best \textsc{ParBalans} vs. Gurobi.”
\end{comment}

\subsubsection{Q2: \textsc{ParBalans} $N$ (\textsc{Gurobi} $T$) vs \textsc{Gurobi} $N \times T$}
In Figure~\ref{fig:q2}, we next investigate the improvements achieved through solver-level parallelism within the \textsc{ParBalans} MIP solver. The configuration pool subsets used in the simulations are sampled from a total of 180 pre-generated configurations. 

\medskip

For instance, in the case of \textsc{ParBalans}(\textsc{Gurobi}-4), simulations draw from a pool of 45 configurations. Similarly, \textsc{ParBalans}(\textsc{Gurobi}-8) uses a pool limited to 20 configurations. These subsets are selected based on the performance of \textsc{ParBalans}(\textsc{Gurobi}-1) across the full set of 180 configurations. Specifically, the top 45, 20 and 10 configurations (ranked by primal gap) are chosen to form the reduced pools. The performance of the \textsc{ParBalans} using these fixed and reduced pools is then compared against \textsc{Gurobi}. 

Figure~\ref{fig:q2} presents a detailed performance comparison between \textsc{ParBalans}-N(\textsc{Gurobi}-T) and Gurobi with $N \times T$ threads across varying levels of parallelization. For \textsc{Gurobi}, performance improvements become most noticeable when the number of threads exceeds 32, for both instance sets. 

For synthetic instances (left side), \textsc{Gurobi}-64 initially outperforms the corresponding \textsc{ParBalans}-8(\textsc{Gurobi}-8). However, as the level of solver parallelization increases to 16, \textsc{ParBalans} begins to outperform parallel Gurobi. For example, \textsc{ParBalans}-2(\textsc{Gurobi}-16) and \textsc{ParBalans}-4(\textsc{Gurobi}-16) outperform \textsc{Gurobi}-32 and \textsc{Gurobi}-64, respectively. 

For real-world instances (right side), \textsc{ParBalans} consistently outperforms parallel \textsc{Gurobi} across all levels of parallelization. Overall, \textsc{ParBalans} demonstrates stronger performance on real-world instances compared to synthetic ones. For a detailed summary statistics, we refer to Appendix B. For performance per problem type, we refer to Appendix C. 

%\todo{For best performing configurations, we refer to Appendix D.}

\section{Related Work}
\label{sec:related}
Incorporating learning-based methods into MIP solving has demonstrated substantial promise in enhancing both exact and heuristic approaches. Within exact methods, particularly branch-and-bound (BnB), this includes learning to branch\cite{BENGIO2021405,khalil2016learning,dash,lodi2017learning,DBLP:conf/aaai/KadiogluMS12,cai2024learning}, and learning to select nodes\cite{NIPS2014_757f843a}. For meta-heuristics, progress includes learning to predict and search for high-quality solutions\cite{kadioglu2017learning,ding2020accelerating,huang2024contrastive,cai2024multi}, integrating ML techniques into large neighborhood search (LNS) frameworks on top of MIP solvers\cite{song2020general,huang2023searching}, and learning to schedule primal heuristics dynamically within the BnB tree~\cite{khalil2017learning,chmiela2021learning,hendel2022adaptive}.

Primal heuristics are essential for quickly finding feasible solutions that tighten the primal bound early in the search. Recently, meta-solver frameworks that embed MIP solvers within LNS have shown superior performance over standard MIP solvers, particularly when combined with learned guidance~\cite{kadioglu2011incorporating,song2020general,tong2024optimization}. A growing body of work explores learning to guide such heuristics. For example, IL-LNS~\cite{sonnerat2021learning} learns to select variables by imitating local branching decisions; RL-LNS~\cite{wu2021learning} adopts a reinforcement learning framework for similar tasks; and CL-LNS~\cite{huang2023searching} uses contrastive learning to predict variables to perturb. Complementing these are online methods such as~\cite{balans}, which adaptively generate high-quality solutions without any offline training, demonstrating competitive performance on difficult instances.

Another line of research focuses on guiding primal heuristic operations within the BnB solver itself. Offline approaches include~\cite{khalil2017learning}, which learns a binary classifier for deciding whether to apply a heuristic at a given node, and~\cite{chmiela2021learning}, which constructs an execution schedule by assigning priorities and compute budgets to different heuristics. In contrast,~\cite{hendel2022adaptive} adopts an online multi-armed bandit approach to dynamically select heuristics during the BnB search.

Parallelization also plays a significant role in improving heuristic performance. Simple parallelization techniques, such as those explored in~\cite{cire2014parallel}, have been shown to be surprisingly effective in accelerating search and enhancing solution quality. However, when running in parallel, ensuring determinism becomes important—especially in production-grade solvers or benchmarking environments—as discussed in~\cite{kadiouglu2024integrating}.

The built-in parallelization capabilities of \textsc{MabWiser}—the bandit-based library used in \textsc{Balans}—have proven successful in diverse applications~\cite{DBLP:conf/aaai/KadiogluK24,textwiser2021,bdl_jsda,bdl_ictai,kadiouglu2024integrating}, and offer a robust framework for deploying online learning in high-performance optimization settings.

\section{Conclusion}   

\begin{comment}

Removed config analysis section for now, since the report would be too long otherwise. There are some interesting results but not statistically strong overall. ( example: simulated annealing in general works great with thompson sampling etc.) I can add one small paragraph about it maybe?

\section{Configuration Analysis}

\begin{figure}[htbp]
    \raggedright
    \includegraphics[width=0.8\textwidth]{figure-5.png}
    \caption{Config similarity}
    \label{fig:Figure 5}
\end{figure}
4) Configuration Analysis
    4.1. Default Balans parameters?
    4.2.All figures if deemed worthwhile. (rest goes to appendix)
\end{comment}

In this paper, we further investigated the parallelization capabilities of the recently proposed BALANS \cite{balans}. Leveraging its highly configurable architecture, we conducted a comprehensive and carefully designed experimental study to evaluate how BALANS scales with parallelization. Our results demonstrate that BALANS benefits from both solver-level and algorithmic-level parallelism, achieving competitive performance compared to state-of-art MIP solver Gurobi across a range of challenging real-world and synthetic MIP benchmark instances.

\printbibliography

%\bibliographystyle{named}
%\bibliography{ijcai25}

\clearpage
\onecolumn
\appendix
\section*{Appendix}
\addcontentsline{toc}{section}{Appendix}

\section{Configuration Space}

\begin{table}[h]
\centering
\small
\begin{tabular}{@{}l l p{8cm}@{}}
\toprule
\textbf{Parameter} & \textbf{Type} & \textbf{Selection Pool} \\
\midrule
Total number of destroy operators & Integer & [4, 16] \\
Destroy percentage & Integer & [10, 20, 30, 40, 50] \\
Destroy operator selection & Categorical & [crossover, mutation, local branching, proximity, rens, rins] \\
Accept criterion & Categorical & [HillClimbing, Simulated Annealing] \\
\quad Step value in Simulated Annealing & Real & [0.01, 1] \\
Learning policy & Categorical & [Epsilon Greedy, Softmax, Thompson Sampling] \\
\quad Epsilon in E-greedy & Real & [0, 0.5] \\
\quad Tau in soft-max & Real & [1, 3] \\
Reward values for [best, better, accept, reject] & Integer & [8,4,2,1], [3,2,1,0], [5,2,1,0], [16,4,2,1], [8,3,1,0], [5,4,2,0], [1,1,1,0], [1,1,0,0] \\
\bottomrule
\end{tabular}
\caption{The configuration space of \textsc{ParBalans}.}
\label{tab:BALANS_parameters}
\end{table}

Table~\ref{tab:BALANS_parameters} lists the complete parameter space used to generate the 180 configurations for our experiments. These configurations are created using a lightweight random sampling algorithm, as depicted in Algorithm 1, where each parameter is drawn from its respective pool. Inter-parameter dependencies and design constraints are also respected. For instance, if a configuration includes \textit{Simulated Annealing} as the acceptance criterion, its associated hyperparameter, the \textit{step value}, is randomly selected from the range $[0.01, 1]$. Similarly, if the \textit{Proximity Search} destroy operator is chosen, the destroy percentage is sampled from the set $[5,10,15,20,30]$ instead. 
\begin{comment}
    is it 5% or 0.5% for proximity. Double check with junyang.
\end{comment}

\newpage
\section{\textsc{ParBalans} vs.\textsc{Gurobi} at Higher Levels of Solver Parallelism}

\begin{table}[H]
\centering
\begin{adjustbox}{max width=\textwidth}
\begin{tabular}{c|c|cc|cc||cc|cc}
\toprule
& \multirow{2}{*}{Degree of Parallelism} 
& \multicolumn{4}{c||}{\textbf{Synthetic}} 
& \multicolumn{4}{c}{\textbf{Real-world}} \\
& & \multicolumn{2}{c|}{\textsc{Gurobi}} & \multicolumn{2}{c||}{\textsc{Parbalans}}
& \multicolumn{2}{c|}{\textsc{Gurobi}} & \multicolumn{2}{c}{\textsc{Parbalans}} \\
\cmidrule(lr){3-4} \cmidrule(lr){5-6} \cmidrule(lr){7-8} \cmidrule(lr){9-10}
& & PG (\%) & PI & PG (\%) & PI & PG (\%) & PI & PG (\%) & PI \\
\midrule
\multirow[c]{6}{*}{\rotatebox[origin=c]{90}{\textbf{pb (Gurobi-8)}}}
& 2   & 2.81±2.74 & 125.97±129.75 & 1.14±1.83 & 54.84±66.65 & 0.50±0.50 & 21.52±21.30 & 0.20±0.21 & 9.51±9.72 \\
& 4   & 0.98±1.12 & 41.75±40.54   & 0.80±1.51 & 41.37±58.22 & 0.37±0.46 & 16.08±18.07 & 0.13±0.14 & 6.96±7.34 \\
& 8   & 0.84±1.08 & 39.04±38.68   & 0.55±1.24 & 31.19±49.52 & 0.38±0.42 & 16.92±17.41 & 0.07±0.08 & 5.00±5.49 \\
& 16  & 0.88±1.54 & 39.05±49.30   & 0.39±1.12 & 23.55±42.12 & 0.42±0.52 & 18.90±19.34 & 0.02±0.05 & 3.34±4.03 \\
& 32  & 0.80±1.62 & 37.28±49.83   & -         & -           & 0.34±0.34 & 15.51±14.57 & -         & -         \\
& 64  & 0.29±0.51 & 21.64±17.49   & -         & -           & 0.15±0.24 & 10.33±11.27 & -         & -         \\
\midrule
\multirow[c]{6}{*}{\rotatebox[origin=c]{90}{\textbf{pb (Gurobi-16)}}}
& 2   & 2.92±2.72 & 128.71±129.13 & 0.64±0.99 & 35.20±40.81 & 0.51±0.51 & 18.01±18.34 & 0.15±0.18 & 7.13±7.97 \\
& 4   & 1.12±1.07 & 46.05±39.64   & 0.36±0.67 & 23.45±29.68 & 0.37±0.46 & 13.08±15.32 & 0.10±0.13 & 5.19±5.94 \\
& 8   & 0.96±1.06 & 42.93±38.36   & 0.11±0.24 & 13.64±17.22 & 0.39±0.43 & 13.88±15.12 & 0.05±0.08 & 3.62±4.21 \\
& 16  & 1.02±1.49 & 43.45±48.28   & -         & -           & 0.42±0.52 & 15.77±17.59 & -         & -         \\
& 32  & 0.92±1.59 & 41.47±48.70   & -         & -           & 0.34±0.36 & 12.64±12.46 & -         & -         \\
& 64  & 0.42±0.54 & 26.47±18.23   & -         & -           & 0.16±0.22 & 7.85±8.76   & -         & -         \\
\bottomrule
\end{tabular}
\end{adjustbox}
\caption{Performance comparison of Parallel Gurobi and \textsc{ParBalans} on synthetic (left) and real-world (right) instances. 
Each entry reports the average Primal Gap (PG, in \%) and Primal Integral (PI) at a 1-hour time limit. 
Lower values indicate better performance. The first subtable (labeled \textbf{pb (Gurobi-8)}) reports results when \textsc{ParBalans} uses Gurobi solver with 8 threads. 
The second subtable (labeled \textbf{pb (Gurobi-16)}) shows results when \textsc{ParBalans} uses Gurobi solver with 16 threads.}
\end{table}

\begin{table}[ht]
\centering
\small
\label{tab:primal_gap_comparison}
\begin{tabular}{c|ccrrr}
\toprule
\textbf{Instance Set} & \textbf{Approach} & \makecell{Degree of\\Parallelism} & \makecell{\textsc{Gurobi}\\Gap (\%)} & \makecell{\textsc{Parbalans}\\Avg. Gap (\%)} & \makecell{\textsc{Parbalans}\\Best Gap (\%)} \\
\midrule

\multirow{14}{*}{Synthetic} 
& \multirow{7}{*}{Pb (Gurobi-8)} 
& 2   & 2.81 & 1.14 & 0.76 \\
& & 4   & 0.98 & 0.80 & 0.54 \\
& & 8   & 0.84 & 0.55 & 0.41 \\
& & 16  & 0.88 & 0.39 & 0.35 \\
& & 32  & 0.80 & --   & --   \\
& & 64  & 0.29 & --   & --   \\
\cmidrule{2-6}
& \multirow{7}{*}{Pb (Gurobi-16)} 
& 2   & 3.03 & 0.64 & 0.29 \\
& & 4   & 1.14 & 0.36 & 0.13 \\
& & 8   & 0.98 & 0.11 & 0.02 \\
& & 16  & 1.01 & --   & --   \\
& & 32  & 0.97 & --   & --   \\
& & 64  & 0.45 & --   & --   \\

\midrule

\multirow{14}{*}{Real-world} 
& \multirow{7}{*}{Pb (Gurobi-8)} 
& 2   & 0.50 & 0.20 & 0.07 \\
& & 4   & 0.37 & 0.13 & 0.02 \\
& & 8   & 0.38 & 0.07 & 0.02 \\
& & 16  & 0.42 & 0.02 & 0.02 \\
& & 32  & 0.34 & --   & --   \\
& & 64  & 0.15 & --   & --   \\
\cmidrule{2-6}
& \multirow{7}{*}{Pb (Gurobi-16)} 
& 2   & 0.51 & 0.15 & 0.09 \\
& & 4   & 0.37 & 0.10 & 0.05 \\
& & 8   & 0.39 & 0.05 & 0.04 \\
& & 16  & 0.42 & --   & --   \\
& & 32  & 0.34 & --   & --   \\
& & 64  & 0.16 & --   & --   \\

\bottomrule
\end{tabular}
\caption{Primal gap comparison across parallelization levels for Gurobi and \textsc{ParBalans} best configuration performance.}
\end{table}

\begin{comment}
the comparison should not be done line by line here. ( eg gurobi 16 corresponds to pb2(gurobi-8) which are not shown in the same row in table.) maybe a better way to design the tables, so that line by line comparison can be done?
\end{comment}

Tables 4 and 5 present the performance of \textsc{ParBalans} compared to (\textsc{Gurobi} when the number of solver threads allocated to (\textsc{Gurobi} within \textsc{ParBalans} is set to 8 and 16, respectively. These tables are analogous to Tables 1 and 2, providing a consistent comparison. Both tables indicate that \textsc{ParBalans} benefits from solver-level parallelism—achieving better performance than (\textsc{Gurobi} across all levels in real-world instances and showing competitive results on synthetic instances. Notably, Table 4 suggests that \textsc{ParBalans}'s flexibility can be fully leveraged through careful configuration design, with the best configuration reducing the primal gap by more than 50\% compared to the average in some cases.

\newpage
\section{Performance Comparison by Problem Type}

The following figures illustrate the performance of \textsc{ParBalans}(\textsc{Gurobi}-1) compared to parallel Gurobi, grouped by synthetic and real-world instance types: General Independent Set Problems (GISP), Maximum Independent Set (MIS), Set Covering (SC), Combinatorial Auctions (CA), Seismic-Resilient Pipe Network Planning problems (SRPN) and Middle-Mile Consolidation Network (MMCN). For ease of comparison, lines representing the same level of parallelism are consistently colored and matched between Gurobi and \textsc{ParBalans} when the parallelism levels are equal.

The figures on the left display the primal gap (\%) for \textsc{Gurobi}, while the figures on the right show the results for \textsc{ParBalans}. Overall, \textsc{ParBalans} demonstrates competitive performance relative to \textsc{Gurobi}. Specifically, for MIS instances, \textsc{Gurobi} performs better at lower levels of parallelism. However, as the degree of parallelism increases to 32, \textsc{ParBalans} begins to outperform \textsc{Gurobi}. For SC and CA problem types, \textsc{ParBalans} achieves comparable performance to \textsc{Gurobi} at higher parallelism levels. In contrast, for GISP instances, \textsc{ParBalans} performs relatively poorly. For SRPN, both solvers reach very small primal gaps quickly, with both achieving a zero gap at 64 threads in a short time. In the MMCN case, \textsc{ParBalans} remains competitive or better up to 32 threads, after which \textsc{Gurobi} slightly outperforms it reaching lower final gap.

\addtocounter{figure}{1}
\begin{figure}[H]
    \centering
    \includegraphics[width=0.75\textwidth]{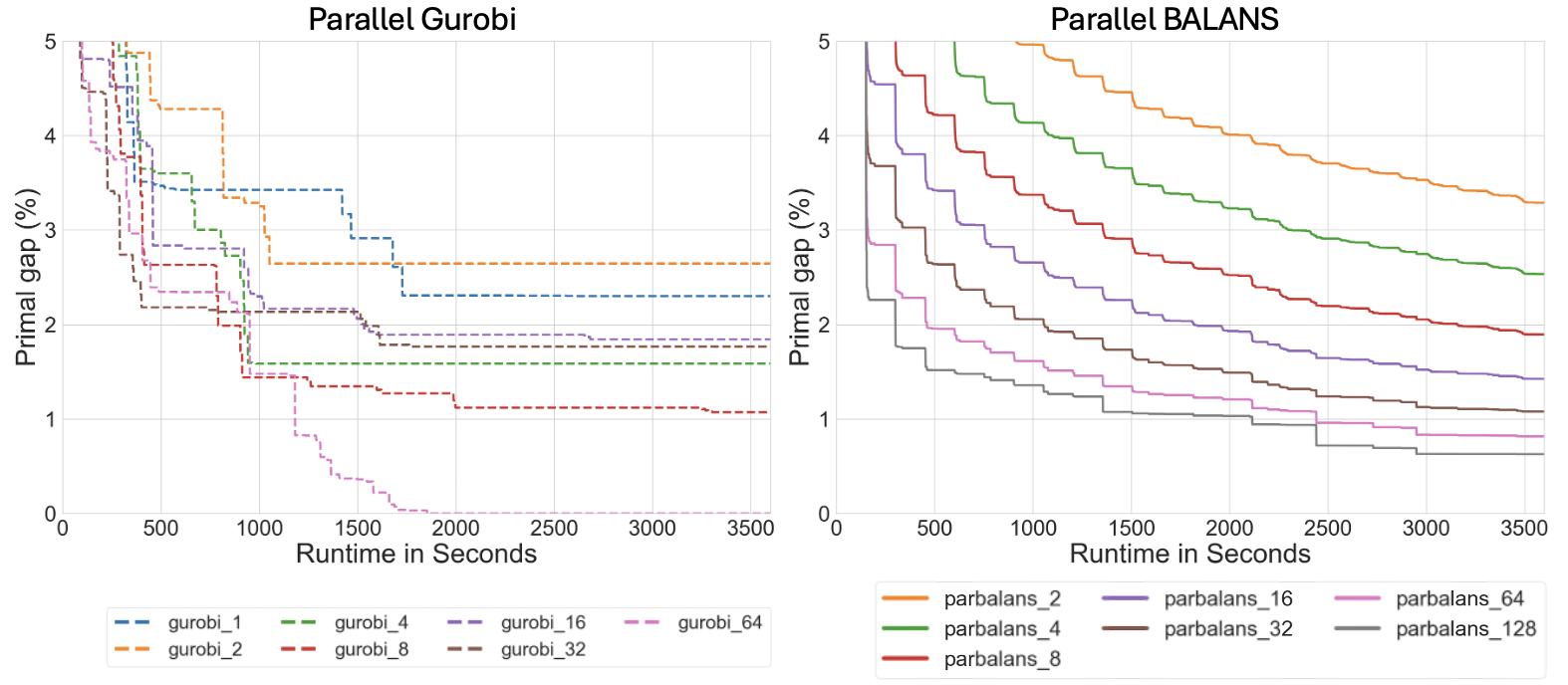}
    \caption{The primal gap (the lower, the better) as a function of time, averaged over instances from \textbf{GISP distribution}.}
    \label{fig:primal_gap_GISP}
\end{figure}
\begin{figure}[H]
    \centering
    \includegraphics[width=0.75\textwidth]{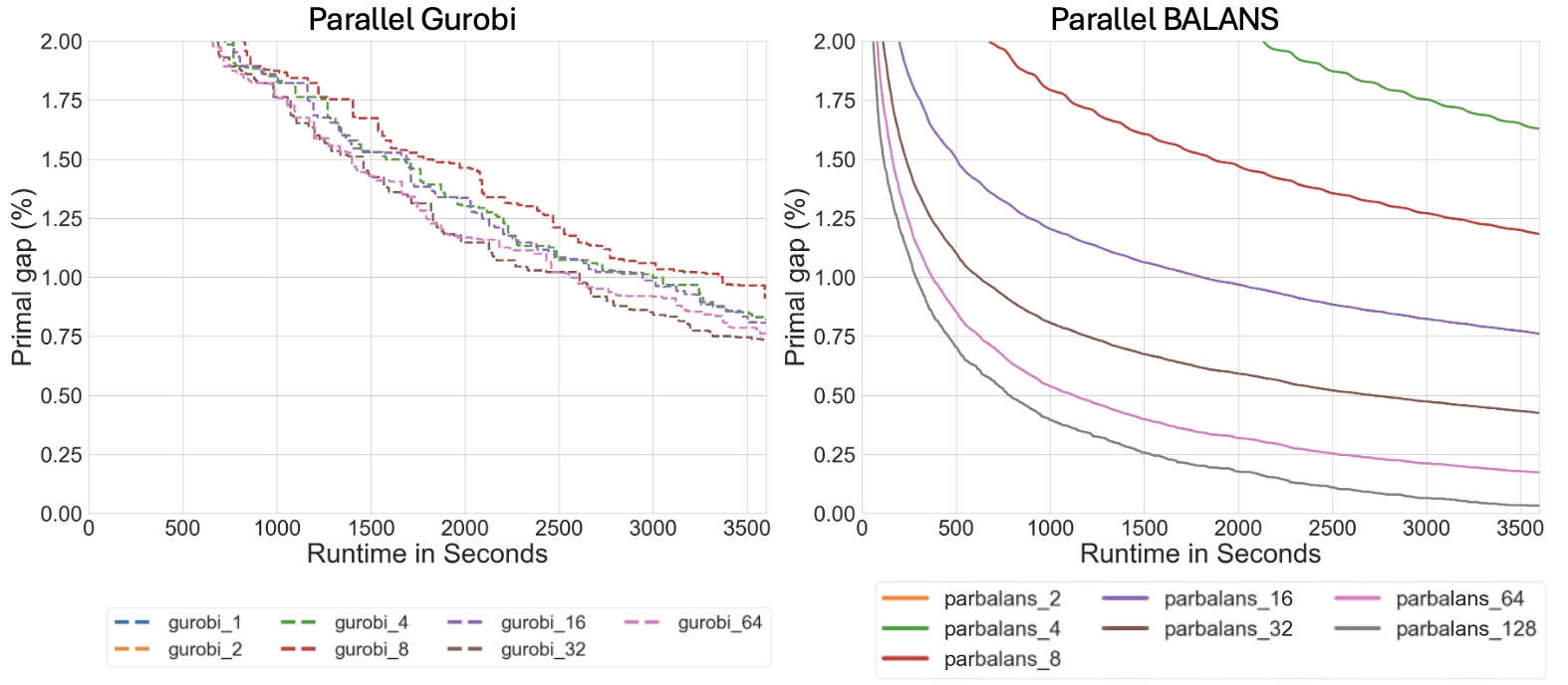}
    \caption{The primal gap (the lower, the better) as a function of time, averaged over instances from\textbf{ MIS distribution}.}
    \label{fig:primal_gap_MIS}
\end{figure}
\begin{figure}[H]
    \centering
    \includegraphics[width=0.75\textwidth]{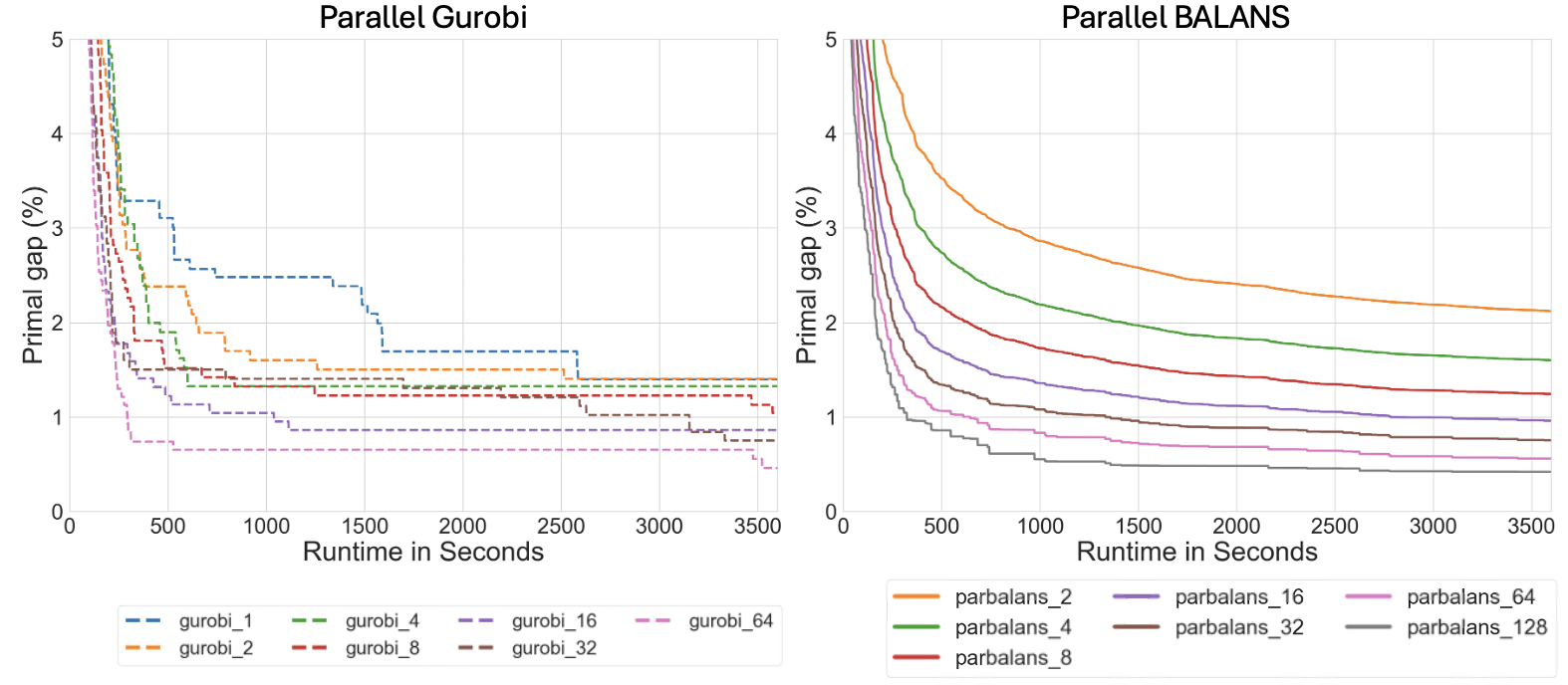}
    \caption{The primal gap (the lower, the better) as a function of time, averaged over instances from \textbf{SC distribution}.}
    \label{fig:primal_gap_SC}
\end{figure}
\begin{figure}[H]
    \centering
    \includegraphics[width=0.75\textwidth]{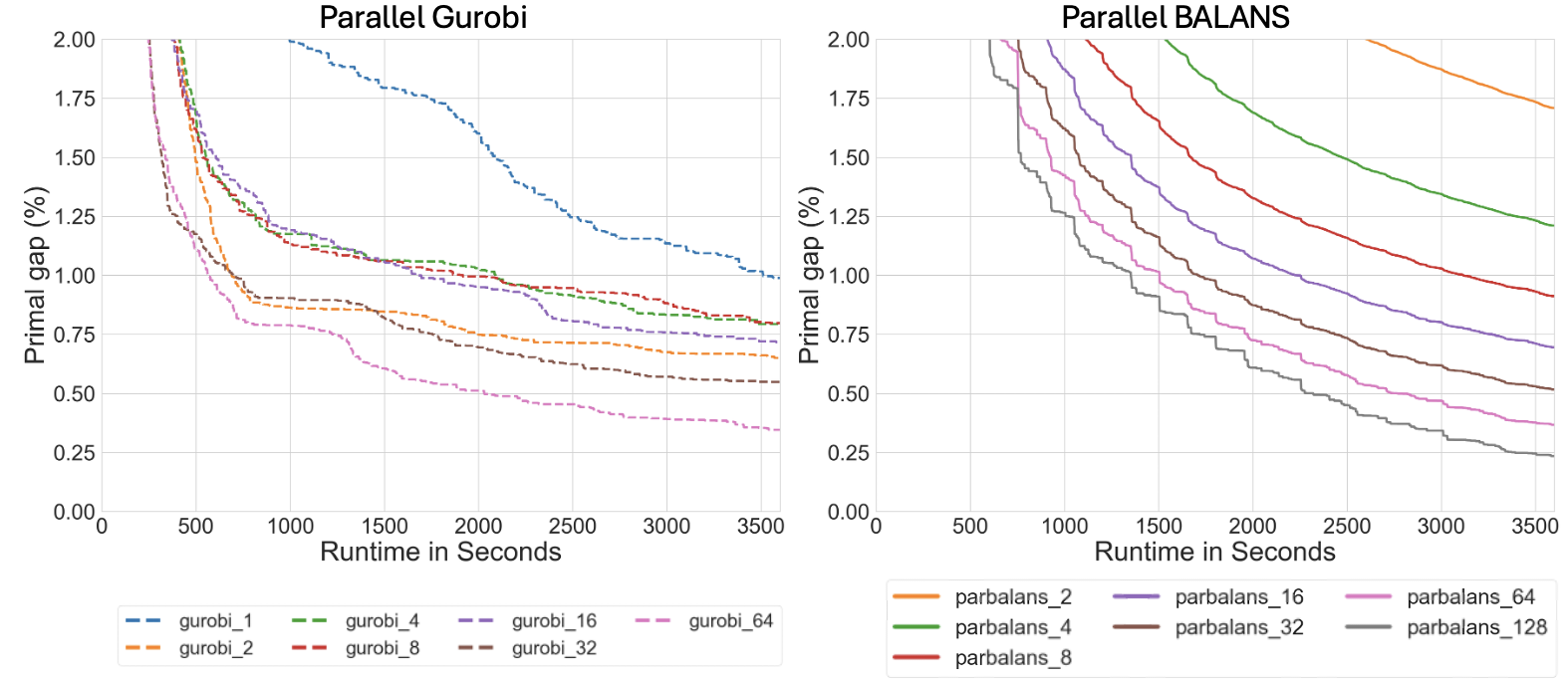}
    \caption{The primal gap (the lower, the better) as a function of time, averaged over instances from \textbf{CA distribution}.}
    \label{fig:primal_gap_ca}
\end{figure}
\begin{figure}[H]
    \centering
    \includegraphics[width=0.75\textwidth]{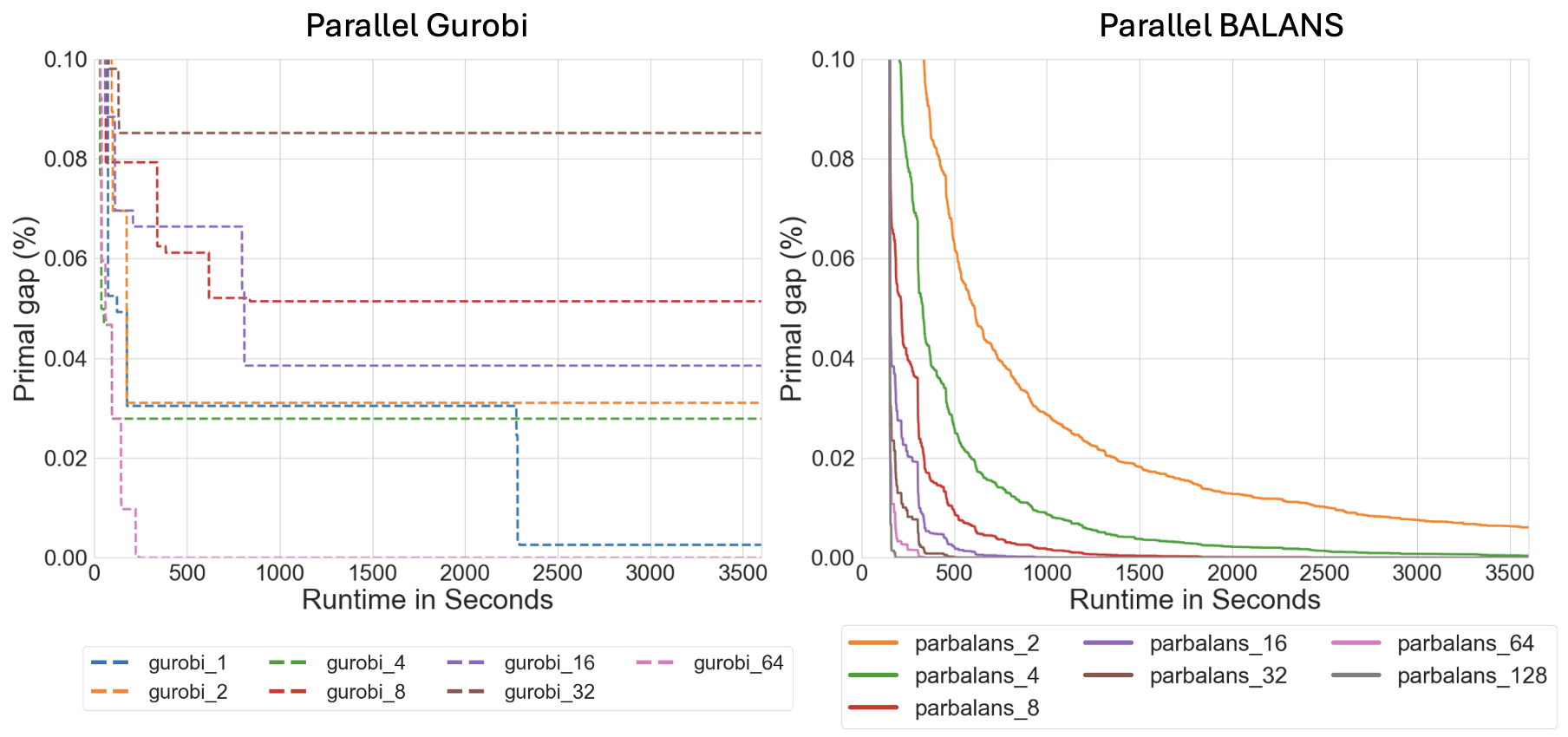}
    \caption{The primal gap (the lower, the better) as a function of time, averaged over instances from \textbf{SRPN distribution}.}
    \label{fig:primal_gap_srpn}
\end{figure}
\begin{figure}[H]
    \centering
    \includegraphics[width=0.75\textwidth]{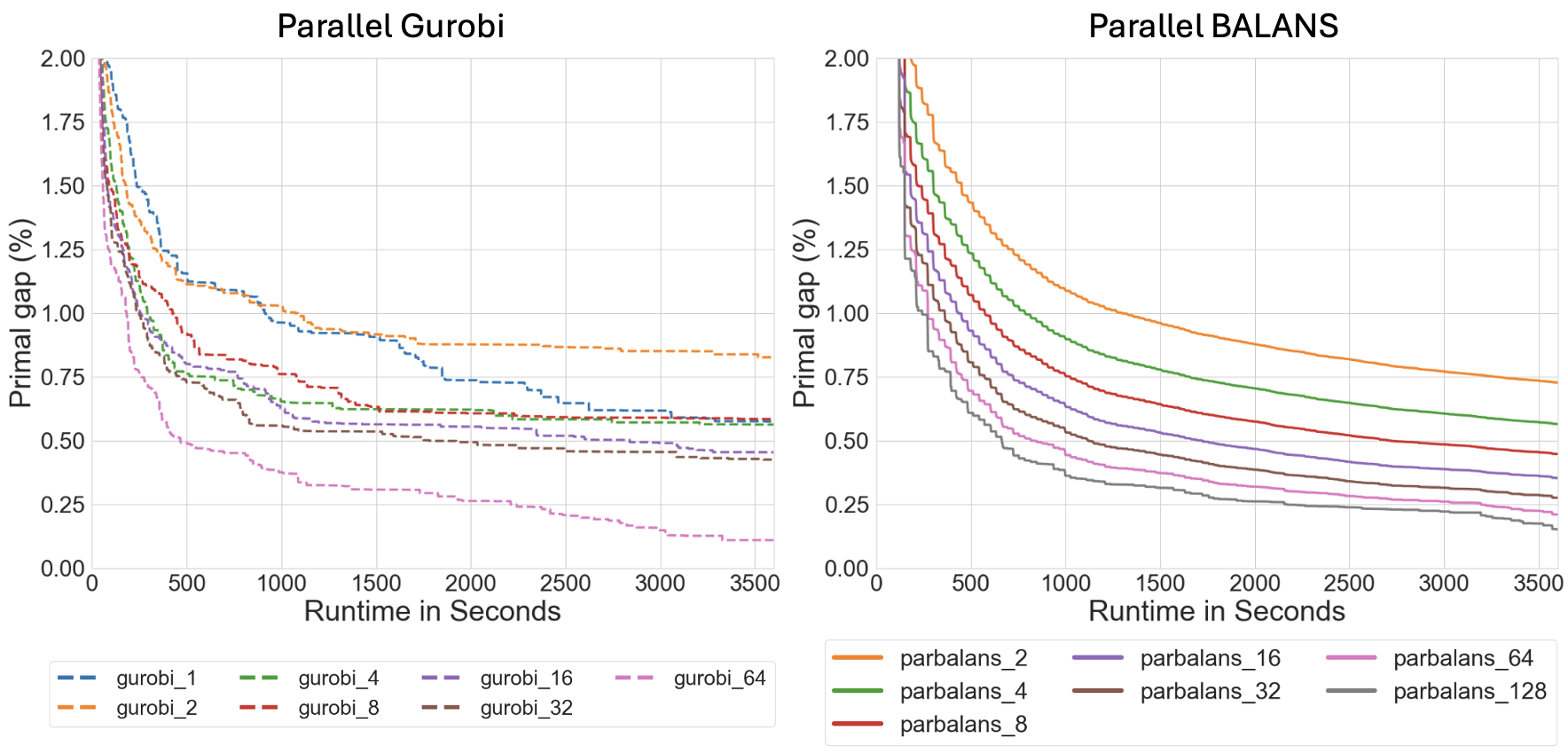}
    \caption{The primal gap (the lower, the better) as a function of time, averaged over instances from \textbf{MMCN distribution}.}
    \label{fig:primal_gap_mmcn}
\end{figure}

% \newpage
% \section{Best Performing \textsc{ParBalans} Configurations}
% \todo{TODO: We report the best performing \textsc{ParBalans} configurations across all instances and per problem type. This is the biggest gap so far in this paper}.

\end{document}